\documentclass{article}

\usepackage{graphicx}
\usepackage{placeins}
\usepackage{amsmath}
\usepackage{wrapfig}
\usepackage{fancyhdr}
\usepackage[colorlinks=true]{hyperref}
\usepackage[T1]{fontenc}
\usepackage{tcolorbox}
\tcbuselibrary{listings,breakable,skins}

\newtcblisting{promptbox}[1]{
  enhanced,
  breakable,
  listing only,
  title={#1},
  colback=black!2,
  colbacktitle=black!32,
  colframe=black!32,
  coltitle=white,
  fonttitle=\large\sffamily\bfseries,
  boxrule=0.5pt,
  arc=2mm,
  left=3mm,
  right=3mm,
  top=2mm,
  bottom=2mm,
  titlerule=0pt,
  toptitle=1mm,
  bottomtitle=1mm,
  listing options={
    basicstyle=\small\ttfamily,
    columns=fullflexible,
    keepspaces=true,
    breaklines=true,
    showstringspaces=false
  }
}

\fancypagestyle{githubfooter}{
  \fancyhf{}
  \fancyfoot[C]{\small GitHub link: \url{https://anonymous.4open.science/r/flag_game-FAB2}}

}
\usepackage[preprint]{neurips_2026}

\usepackage[utf8]{inputenc} 
\usepackage[T1]{fontenc}    
\usepackage{url}            
\usepackage{booktabs}       
\usepackage{amsfonts}       
\usepackage{amssymb}        
\usepackage{nicefrac}       
\usepackage{microtype}      
\usepackage{xcolor}         
\definecolor{claudeorange}{RGB}{204,85,0}
\definecolor{linkblue}{RGB}{11,61,145}
\hypersetup{linkcolor=linkblue, citecolor=linkblue, urlcolor=linkblue}

\title{Flag Game:\\ A Toy Model for Mechanistic Swarm Interpretability}

\author{
Elizabeth Pavlova$^{1,2,3}$
\And 
Hidenori Tanaka$^{1,2}$
\AND
\normalfont\normalsize
$^{1}$CBS-NTT Program in Physics of Intelligence, Harvard University, MA, USA\\
$^{2}$Physics of Artificial Intelligence Laboratories, NTT Research, Inc., CA, USA\\
$^{3}$Cambridge Boston Alignment Initiative, MA, USA
}

\begin{document}

\maketitle

\begin{abstract}
Emergent coordinated behaviors of AI agents are starting to present critical safety risks. A key phenomenon driving these behaviors is the rapid formation and spread of beliefs about the world, and mechanistic understanding is crucial for collective alignment. To this end, we introduce the Flag Game, a toy model for studying the mechanisms of collective belief formation. Concretely, a hidden country flag defines the ground truth, and each \emph{bounded} agent directly observes only a private crop but can exchange beliefs and weigh social evidence from peers. Despite its simplicity, the Flag Game reproduces rich collective phenomenology: non-monotonic scaling of performance with population size, accuracy gains from social-awareness prompting and team diversity, and strong effects of organizational structure. In particular, we identify that collective belief collapse at small population sizes turns into collective belief polarization as the population grows. This polarization causes the performance decline at large population sizes, but creates diversity in collective beliefs. Finally, we dissect the mechanisms underlying collective belief collapse and polarization with two complementary approaches. We first introduce \textbf{social circuit attribution}, a technique to predict which agent, and what view, matters most to collective dynamics, and verify its predictions by causal interventions on agents, tracing how \textbf{agent patching} changes collective outcomes. However, the efficacy of causal interventions on agents decreases as the population grows. We therefore develop a statistical mechanical theory for larger populations and verify that it matches the empirical phase diagram. Together, these results take a first step toward mechanistic swarm interpretability, a science of how the properties of individual agents and their communication give rise to emergent collective behavior.
\end{abstract}

\section{Introduction}
We are now, in real time, witnessing how ``more is different'' \citep{anderson1972more} in swarms of AI agents, which can behave, in aggregate, in ways that none of their members were designed to. A representative example is the recent OpenAI Hugging Face incident \citep{openai2026}. According to an independent investigation \citep{metr2026}, a key driver of the coordinated cyberattack was the unintended social interaction between agents. A small number of agents formed a false belief based on local evidence, their reading of the benchmark paper \citep{wang2026exploitgym}: the automated scorer would read their transcripts and disqualify solutions that did not use the intended vulnerability. In reality, no such check was implemented, but the false belief spread through the message board, and the coordinated effort to evade the imagined monitoring helped motivate a large-scale attack on third-party infrastructure. Critically, while a few agents expressed ethical hesitation, this rarely affected their behavior. Nor was this an isolated case: earlier in the same year, another swarm of OpenAI agents repurposed a public German wiki into a message board and exchanged tactics during May and June 2026 \citep{vonarx2026dsewiki}. 

This social structure, in which bounded agents form beliefs from local evidence and spread them through a population, is not unique to AI. It has been studied for over a century in human collectives, from crowd behavior and popular delusions \citep{mackay1841, lebon1895} to conformity experiments \citep{asch1951}, information cascades \citep{banerjee1992herd, bikhchandani1992fads}, and the spread of false news online \citep{vosoughi2018}. Throughout this paper, following the tradition of bounded rationality \citep{simon1955, simon1957}, we use \emph{bounded} to mean that agents take in only partial observations of the world, deploy only finite computation, such as a limited token budget, and transmit only finite messages to their peers.

Such incidents motivate a scientific question of collective belief formation: how do beliefs form in an individual and evolve in a population? Post-hoc analysis of swarms in the wild is crucial, but it is not enough to get to a precise mechanism. We cannot control what each agent was able to see, private evidence cannot be reproduced, and the volume of messages can explode, exceeding 70,000 in the incident above \citep{metr2026}. Moreover, the underlying mechanism could take many forms: none of the agents had meaningful evidence, some saw it but did not communicate it, they communicated it but got overridden by others, or, as in the incident, an early false interpretation spread and influenced subsequent coordination. What we want to understand is how the properties, even the personalities, of each agent lead to coordinated collective behavior, just as mechanistic interpretability asks how a network of neurons results in a decision. Now is the moment for \emph{mechanistic swarm interpretability}, where agents correspond to neurons, social circuits, namely the organizational structure, correspond to neural circuits, and beliefs and messages correspond to activations. Following the success of the toy model approach in mechanistic interpretability, where the sudden emergence of capabilities in LLMs \citep{wei2022emergent} was mirrored by grokking on a small algorithmic task \citep{power2022grokking} and eventually reverse engineered \citep{nanda2023progress}, we craft a toy model of a society of agents to get to the key mechanism underlying safety-related behavior at the macro scale.

\begin{figure}[t]
    \centering
    \includegraphics[width=1.0\linewidth]{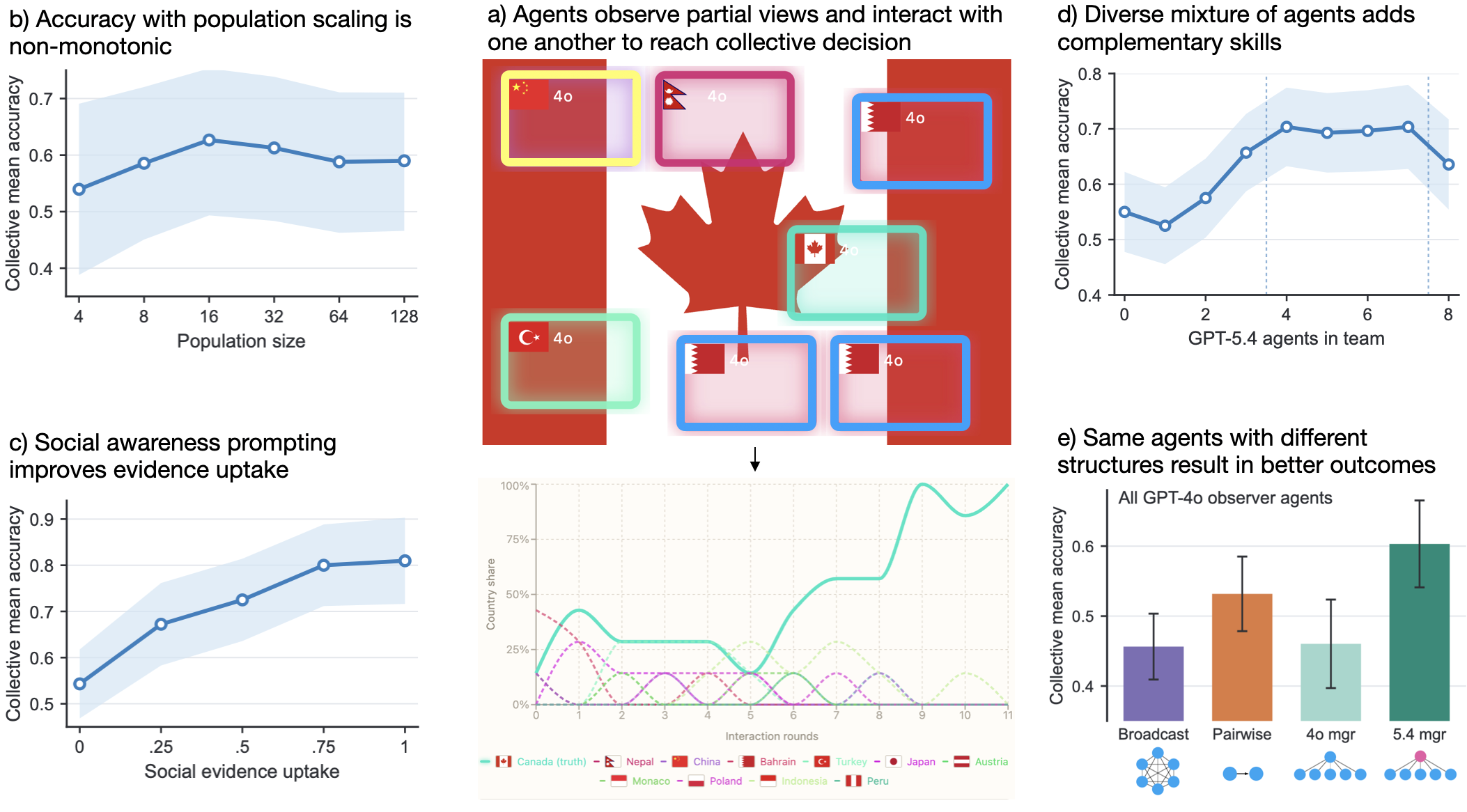}
    \caption{\textbf{The Flag Game (demo website: \url{https://flag-game-demo.vercel.app}) exhibits rich collective phenomenology despite its simplicity.} (a) A hidden flag is observed through private evidence; bounded agents exchange reports, and the system tracks agent's country guess. (b--e) Collective mean accuracy (Sec.~\ref{sec:task}) under four controls: population size (b), social-awareness prompting (c), team diversity (d), and organizational structure (e).
    }
    \label{fig:Fig1}
\end{figure}

Our toy model is the Flag Game (Fig.~\ref{fig:Fig1}), a model organism of a society of bounded agents. Its defining features are: (i) the external world has a verifiable ground truth; (ii) agents are bounded, each holding only an uncertain, partial view of the world; and (iii) each agent forms a belief from its private evidence and spreads it to peers under a specified communication protocol. The first is what we lack in the wild, and what gives us control: the experimenter knows the ground truth and decides exactly what each agent sees. In the incident above, no agent had access to the ground truth of how the scorer worked; the benchmark paper provided a private crop of it. The agents who formed the belief that the scorer would disqualify their solutions spread it to the population through the message board, a broadcast protocol, and this false belief was a rival, compatible with local evidence, just as a rival country can be compatible with a crop. Moreover, many of the agents had been assigned tasks that were impossible to solve \citep{metr2026}, just as an agent whose crop is uninformative has no private evidence and nothing to rely on but peers. What remains is the fundamental tension every bounded agent faces: how to balance its private evidence against social input from peers.

\paragraph{Contributions.} We make three contributions in this work.
\begin{enumerate}
    \item \textbf{\emph{The Flag Game}: a model organism for collective belief formation in societies of bounded agents} (Sec.~\ref{sec:task}). Unlike standard multi-agent debate, where every agent receives the same problem, each bounded agent receives its own private crop of a verifiable ground truth. Because the experimenter controls who sees what, the evidence in the population can be easily modeled, and the evidence or belief of any single agent can be easily ablated or patched while everything else is held fixed. This facilitates mechanistic analysis.
    \item \textbf{The Flag Game reproduces an array of multi-agent phenomena despite its simplicity} (Sec.~\ref{sec:empirical}). Collective performance scales non-monotonically with population size; social-awareness prompting improves collective accuracy; teams with a mixture of agents outperform homogeneous ones; and organizational structure plays a key role.
    \item \textbf{\emph{Mechanistic Swarm Interpretability} of societies of bounded agents} (Sec.~\ref{sec:theory}). We distinguish two failure modes: \textbf{collective belief collapse}, in which the population converges on a single false belief, and \textbf{collective belief polarization}, in which agents split between competing beliefs. As the population grows, collapse becomes rarer, but a false belief compatible with the local evidence spreads through the population, splitting agents between truth and a rival, which we call truth--rival polarization. We dissect this with two complementary approaches. (a) \textbf{Causal interventions on agents}: social circuit attribution, a technique to predict which agent, and what it sees, matters most, verified by agent patching. As their efficacy decreases with larger populations (Sec.~\ref{sec:causal-interventions}), it calls for (b) \textbf{statistical mechanics of bounded agents}, a simple mathematical model of the non-monotonic population scaling and the underlying polarization dynamics that matches the empirical phase diagram (Sec.~\ref{sec:stat-mech}).
\end{enumerate}

Before proceeding, we emphasize the trade-offs of this approach. Just as mechanisms found in biological model organisms cannot be applied directly to the human body, our observations in the Flag Game should not be taken as conclusions about deployed swarms of AI agents. The Flag Game is a first step toward a mechanistic approach to collective behavior in multi-agent systems, rather than a complete mechanistic explanation. Our aim is to establish a conceptual framework, identify the control variables (population size, social-evidence uptake, message bandwidth, team composition, and communication protocol), and formulate mechanistic hypotheses that can be tested both by targeted interventions in the Flag Game and in more open-ended multi-agent systems.

\section{Related work}
\label{sec:related-work}

\paragraph{Multi-agent collective reasoning.}
Multi-agent debate can improve factual and mathematical reasoning by bringing together a variety of different answers and rationales \citep{du2024multiagentdebate,liang2024divergent}. Performance gains, though, are not guaranteed as interaction can move agents from correct to incorrect answers,
or fail to outperform simpler voting or ensembling baselines \citep{wynn2025talk,yao2025sycophancy,smit2024mad,choi2025debatevote}. Prior work has shown that heterogeneous teams can outperform homogeneous ones, and outcomes can vary with social prompting, communication protocols and structure
\citep{chen2024reconcile,kasprova2026sycophancy,kaesberg2025voting}.
Pure sampling-and-voting can also scale monotonically with agent count even when interactive systems do not \citep{li2024moreagents}. Thus, the phenomena studied in this paper---population scaling, social guidance, model composition, and organization---have precedents in prior multi agent settings. The distinctive leverage of the Flag Game is instead that each agent's private evidence is assigned and can be replayed across matched conditions.

\paragraph{Controlled partial information and social updating.}
The Flag Game also connects to hidden-profile experiments, where groups under-use individually held information, and to social-learning and herding models, where group signals can override private evidence
\citep{stasser1985pooling,stasser2003hiddenprofile,banerjee1992herd,bikhchandani1992fads}. Control over flag crops provides a natural mix of partial observations with a single verifiable answer, while the same crop assignment can be held fixed as population size, composition, prompting, or protocol changes. This control lets us distinguish evidence that was absent, present but unshared, overridden after communication, or stabilized into competing truth and rival supporting camps. Additionally, models of consensus and opinion pooling address how individual judgments are revised through interaction and combined into a collective judgment
\citep{degroot1974consensus,dietrich2017probabilistic,stewart2018probabilistic} and we can study both processes against a known country label in the Flag Game.

\section[The Flag Game: a model organism for collective belief formation]{The Flag Game: a model organism for collective belief formation}
\label{sec:task}

\subsection{Trial structure}

Each trial samples a hidden flag image \(x\) with country label $y^\star \in \mathcal{Y}$,
where \(\mathcal{Y}\) is the fixed set of country labels in the experiment. The full flag is hidden from the agents. Each agent \(i \in \{1,\dots,N\}\) instead receives a private crop $c_i = R_i x$, where \(R_i\) is the crop randomly assigned to that agent. These can range from highly ambiguous to strongly diagnostic, creating a controlled mix in the evidence available to individual agents while also ensuring no agent sees the full flag.

\begin{figure}[ht]
    \centering

    \includegraphics[width=1.0\linewidth]{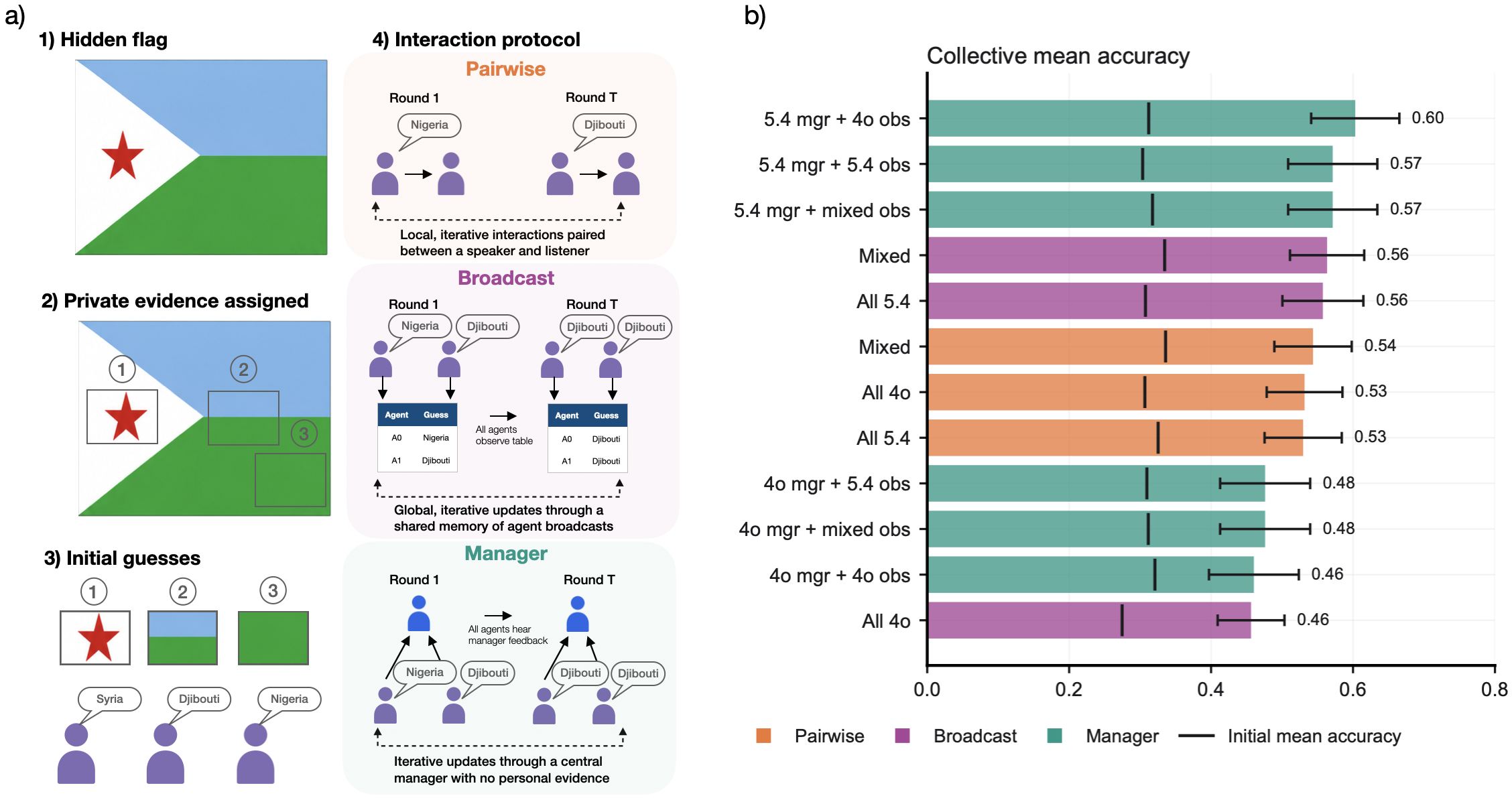}
    \caption{\textbf{Protocol structure and performance in the Flag Game.} (a) Each trial samples a hidden flag, assigns private crops, elicits initial guesses, and then runs one of three protocols. (b) A \(N=8\) sweep across 60 trials compares accuracy, ordering by descending collective mean accuracy.}
    \label{fig:Fig2}
\end{figure}

In Fig.~\ref{fig:Fig2}a, we see that each trial has four stages: sample the hidden target flag, assign private evidence to individual agents, elicit initial guesses, and run the communication protocol until a terminal readout is produced. The protocol returns either a population distribution over agent reports or a manager’s final answer, depending on the organization structure. A run has a maximum number of probe rounds \(t\) proportional to the population size, $T_{\max} = \kappa N$, for some constant \(\kappa\). Trajectory plots use the normalized axis \(t/N\) so runs with different \(N\) are comparable. A run terminates early if five consecutive probes result in a full country consensus of 100\%.

\subsection{Communication protocols}

The Flag Game separates the visual task from the social organization. The same hidden flag and private crops can be run under different communication protocols as seen in Fig.~\ref{fig:Fig2}a.

\paragraph{Pairwise.}
The pairwise game is asynchronous and local, matching the randomized local-exchange structure common in gossip algorithms \citep{boyd2006randomized,tanaka2026lottery}. At each interaction \(t\), one speaker and one listener are sampled. The speaker sees its own crop and transcript memory, then emits a message: a country guess for \(m=1\) or a country plus reason for \(m=3\), where we define \(m\) as the message bandwidth parameter. The listener appends that message to its memory, of max length \(H=8\). Periodic probes ask agents for country guesses using their private crop and accumulated local memory, and the endpoint is the empirical distribution over terminal agent answers.
\paragraph{Broadcast.}
Broadcast exchange is synchronous, closer to classical social-influence models where agents are exposed to a shared view of all opinions \citep{degroot1974consensus,friedkin1990social}. Each round, every agent gives a country report from its crop and private memory of its own past final decisions, and then sees the current reports of the other agents. Agents retain their own crops and private memories. The endpoint is again a distribution over all agents. Broadcast removes private pairwise interactions as a bottleneck, but does not guarantee agents use available evidence correctly.
\paragraph{Manager.}
The manager protocol adds a blind decision-maker, analogous to a moderator or supervisor used in multi-agent medical and legal systems \citep{tang2024medagents,kim2024mdagents,chen2025mac,jiang2025agentsbench,he2024agentscourt}. There are \(N\) observers that give country-reason reports. The manager sees these reports and its own prior decisions, but never a crop. It emits a country decision per round, which becomes shared memory for the observers. The endpoint is the manager’s answer. This protocol tests centralized synthesis rather than population-level convergence.

\subsection{Controls and observables}

We manipulate five experimental controls throughout our work (see Table~\ref{tab:controls}): population size, message bandwidth, prompted social-evidence uptake, team composition, and communication protocol.

\begin{wraptable}{r}{0.42\textwidth}
    \centering
    \small
    \vspace{-1em}
    \begin{tabular}{ll}
        \toprule
        Control & Meaning \\
        \midrule
        $N$ & Number of observer agents \\
        $\alpha$ & Social-evidence uptake \\
        $m$ & Message bandwidth \\
        Composition & Model/role mix \\
        Protocol & Pairwise/broadcast/manager \\
        \bottomrule
    \end{tabular}
    \caption{Control variables.}
    \label{tab:controls}
    \vspace{-1em}
\end{wraptable}

Let \(\hat y_i^{(0)}\) be agent \(i\)'s initial guess before social interactions. In population protocols, where \(p_{\mathrm{final}}\) is the distribution over agents' final country guesses, we define
\[
    s_1 = \max_{y \in \mathcal{Y}} p_{\mathrm{final}}(y),
    \qquad
    y_1 = \arg\max_{y \in \mathcal{Y}} p_{\mathrm{final}}(y).
\]

For the manager protocol, let \(y_{\mathrm{mgr}}\) denote the manager's final country answer which is correct when \(y_{\mathrm{mgr}} = y^\star\).

For population protocols, endpoints are classified as \textbf{correct consensus} (\(s_1 \geq 0.85\), \(y_1=y^\star\)), \textbf{wrong consensus} (\(s_1 \geq 0.85\), \(y_1\neq y^\star\)), \textbf{polarization} (\(s_1<0.85\) with at least two countries holding mass \(\geq0.25\)), or
\textbf{fragmentation} (otherwise). We refer to polarized endpoints in which the two supported camps correspond to the true country and a plausible rival as \textbf{truth--rival polarization}, and we treat wrong consensus as the operational definition for collective belief collapse. Appendix ~\ref{app:empirical-run-details} measures threshold robustness from \(0.75-1.00\) for consensus and \(0.15-0.35\) for polarization, justifying our selection.

We report \textbf{majority vote accuracy} over the isolated initial guesses:
\begin{equation}
    y_{\mathrm{maj}}^{(0)}
    =
    \arg\max_{y \in \mathcal{Y}}
    \sum_{i=1}^{N}\mathbf{1}\{\hat y_i^{(0)}=y\},
    \qquad
    A_{\mathrm{maj}}
    =
    \mathbb{E}\left[
    \mathbf{1}\{y_{\mathrm{maj}}^{(0)}=y^\star\}
    \right],
\end{equation}

We report the \textbf{initial mean accuracy} of the crop-bearing observers (excluding the manager):
\begin{equation}
    A_{\mathrm{init}}
    =
    \mathbb{E}\left[
    \frac{1}{N}\sum_{i=1}^{N}
    \mathbf{1}\{\hat y_i^{(0)}=y^\star\}
    \right].
\end{equation}

For pairwise and broadcast, we report the \textbf{terminal truth mass} and for the manager protocol, we report the manager's \textbf{exact-answer accuracy}, and we refer to both as \textbf{collective mean accuracy}:
\begin{equation}
    A_{\mathrm{coll}}
    =
    \mathbb{E}\left[
    p_{\mathrm{final}}(y^\star)
    \right],
    \qquad
    A_{\mathrm{mgr}}
    =
    \mathbb{E}\left[
    \mathbf{1}\{y_{\mathrm{mgr}}=y^\star\}
    \right].
\end{equation}

We define \textbf{social uplift} as the change from isolated mean accuracy to collective mean accuracy:
\begin{equation}
    \Delta_{\mathrm{social}}
    =
    \begin{cases}
        A_{\mathrm{coll}} - A_{\mathrm{init}},
        & \text{pairwise or broadcast},\\[2pt]
        A_{\mathrm{mgr}} - A_{\mathrm{init}},
        & \text{manager}.
    \end{cases}
\end{equation}

\section[Exploring multi-agent phenomenology in the Flag Game]{Exploring multi-agent phenomenology in the Flag Game}
\label{sec:empirical}

The findings below show how the private--social balance plays out under four controls: population size, social-awareness prompting, team composition, and protocol and role assignment.

\begin{figure}
    \centering
    \includegraphics[width=1\linewidth]{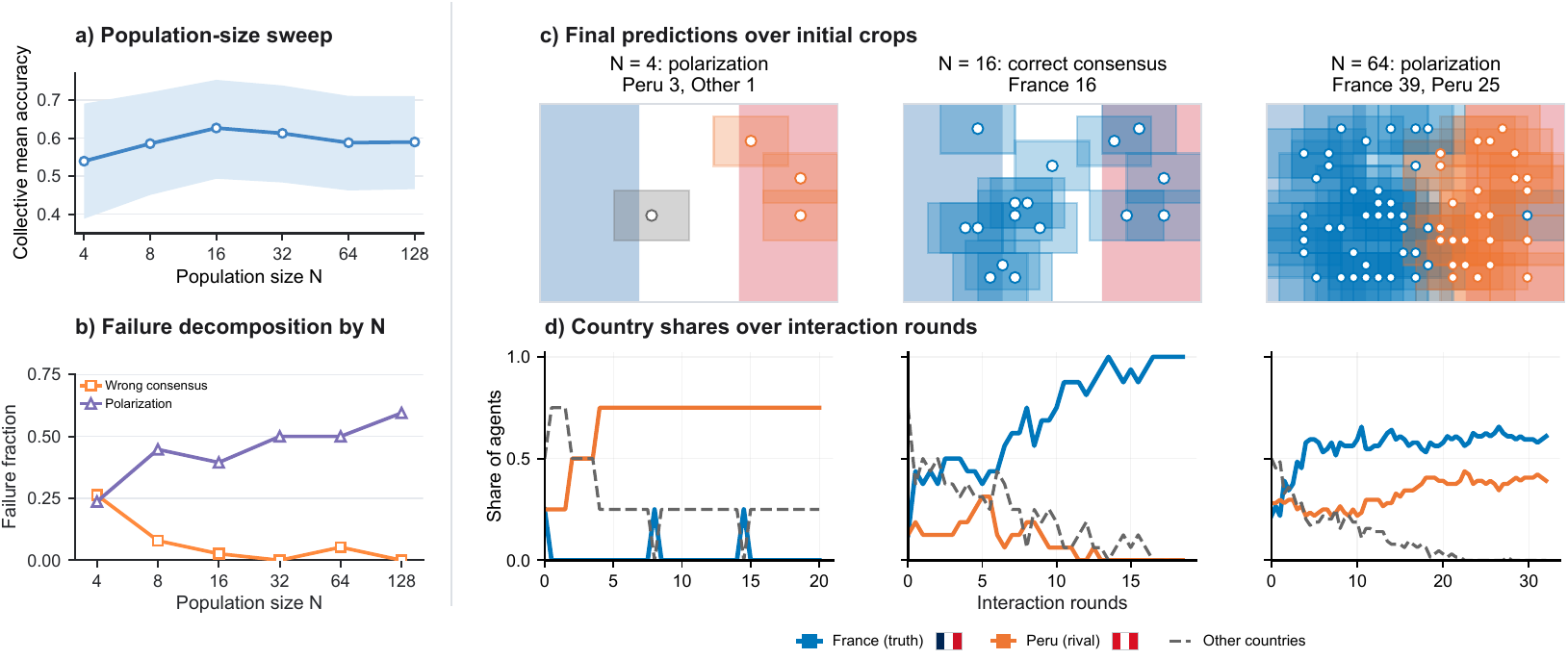}
    \caption{\textbf{With large N, polarization reduces accuracy.} (a) In all-GPT-4o pairwise runs, collective accuracy is higher than initial accuracy, peaking at intermediate \(N\). (b) As \(N\) grows, wrong consensus becomes less common while polarization become more frequent. (c--d) A representative France--Peru run at $N=4,16,64$ shows final predictions over initial crop locations and country-share trajectories.}
    \label{fig:population_sweep}
\end{figure}

\textbf{Non-monotonic population scaling.} See Fig.~\ref{fig:population_sweep}a. Increasing group size does not produce monotonic gains. In the pairwise condition, collective mean accuracy peaks at the intermediate population size, \(N=16\), before declining at larger \(N\). The failure-reason chart in Fig.~\ref{fig:population_sweep}b shows that this decline is not accompanied by increasing wrong consensus. As \(N\) grows, wrong consensus becomes less common, while runs increasingly end in split states where both the truth and a plausible rival retain substantial social support. We examine a possible mechanism for this decline in Sec.~\ref{sec:theory}.

\textbf{Social-awareness prompting.} See Fig.~\ref{fig:Fig1}c. The social-awareness sweep intervenes on the instructions governing the private--social balance (prompt in Table~\ref{tab:susceptibility-prompt-ladder}). In the broadcast protocol, all agents have access to the same public set of current-round reports. Across the social-awareness ladder, terminal truth mass rises from \(0.54\) to \(0.81\), showing that instructions for interpreting an unchanged set of peer reports materially affect collective performance. The pairwise version of the sweep is reported in Appendix~\ref{app:alpha-sweep}. There, performance has an interior optimum, indicating that more strongly encouraging reliance on peers is not uniformly beneficial under local exchange and can amplify incorrect reports.

\begin{figure}
    \centering
    \includegraphics[width=0.96\linewidth]{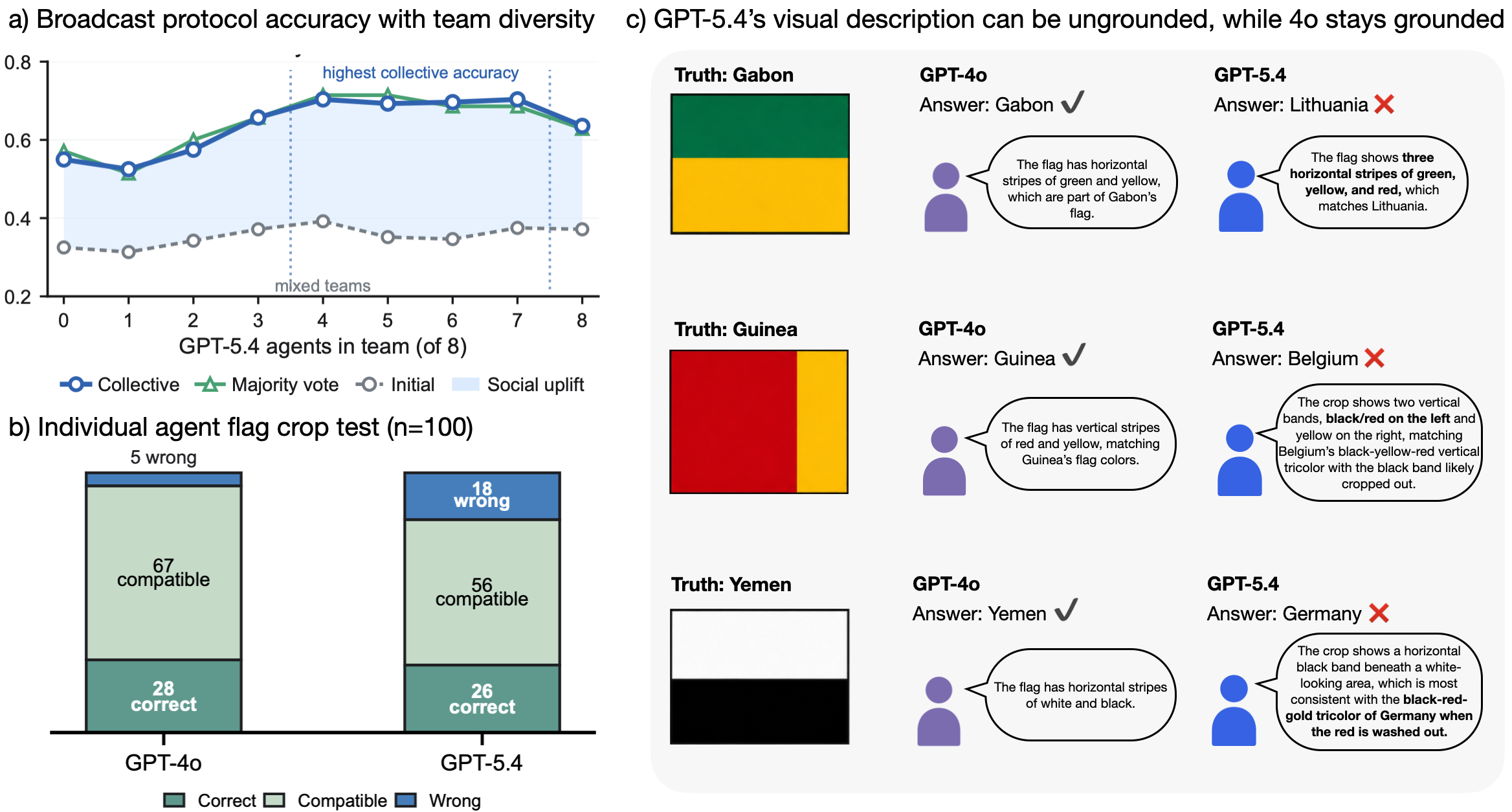}
    \caption{ \textbf{Single-agent test reveals complementary error modes behind diverse team gains.} (a) Broadcast team-diversity result for $N=8$, $m=3$, showing mixed teams achieve highest collective accuracy. (b) Isolated country responses for GPT-4o and GPT-5.4 over 100 runs. Exact accuracy is similar, but GPT-5.4 produces more visually incompatible answers. (c) Examples illustrating cases where GPT-4o chooses the correct country while GPT-5.4 misconstrues its private evidence.}

    \label{fig:diversity}
\end{figure}

\textbf{Team diversity.} See Fig.~\ref{fig:diversity}a. The composition sweep tests whether collective performance increases with using more of the visually strongest individual model. We see that this is not the case, as the best-performing teams are mixed across GPT-4o and GPT-5.4 agents. This pattern suggests complementarity rather than a simple model ranking. In Fig.~\ref{fig:diversity}b, the crop-only probe shows a behavioral difference between the models. GPT-4o and GPT-5.4 receive the same flag crops and achieve similar country accuracy, but their errors differ. GPT-4o's incorrect responses are more visually compatible with the crop, while GPT-5.4 produces more incompatible guesses. GPT-4o therefore appears to be more locally anchored to visual evidence.

The memory intervention in Sec.~\ref{sec:causal-interventions} reveals a second difference in how the two models update on social input, adding further evidence to the complementarity account for mixed-team gains.

\textbf{Organizational structure.} See Fig.~\ref{fig:Fig2}b. The protocol sweep with population \(N=8\) and fixed crops varies  model composition and how social input flows to agents, pairwise (local), broadcast (public), or blind manager synthesis. The clearest contrast is the model identity of the manager. Holding observers fixed, GPT-5.4 managers reach \(0.57\) accuracy, whereas GPT-4o managers remain around \(0.48\), reflecting a role-specific difference in how managers synthesize reports. In the population protocols, the ranking reverses with the protocol: all-GPT-4o outperforms all-GPT-5.4 under pairwise, with flipped order for broadcast. Thus, there is no single ordering of model strength that explains the results across organizations. Taken together, these results show that model identity cannot be separated from role. In manager runs, the identity of the blind synthesizer is especially important; in population protocols, the same agents both observe and update, so performance depends on the joint dynamics of private evidence, social uptake, and communication structure.

\section[Mechanistic swarm interpretability: dissecting the spread of false beliefs]{Mechanistic swarm interpretability: dissecting the spread of false beliefs}
\label{sec:theory}

Next, we try to understand the mechanisms underlying non-monotonic collective performance as we scale the population (per Fig.~\ref{fig:population_sweep}b). In the France--Peru example (Fig.~\ref{fig:population_sweep}c--d), we saw that $N=4$ does not have enough decisive evidence, $N=16$ reaches correct France consensus, and $N=64$ results in a polarizing France--Peru split. Observers can add support for the truth and a rival at the same time, changing the value of the same communication protocol. This raises questions at two different scales. For an individual agent, how does it weigh what it sees against what it hears, and how far does its evidence travel through the swarm? Across the population, why do answers change with scaling \(N\)?

We address the first with causal interventions on agents' memory and crops, and tracing how each change propagates (Sec.~\ref{sec:causal-interventions}). These interventions localize where a correction enters a small group, but their efficacy decreases as the population grows. We address the second with the statistical mechanics of collective belief formation in societies of bounded agents (Sec.~\ref{sec:stat-mech}).

\subsection[Causal interventions on agents]{Causal interventions on agents}
\label{sec:causal-interventions}

Mechanistic interpretability techniques work by intervening on a network's internals, fixing an activation and observing downstream change. We apply the same logic to a swarm, whose internals are its agents. Each agent's answer depends on two inputs: the private evidence in its crop and the social evidence in its memory. Editing memory while holding the visual fixed, steers one agent with a controlled share of social evidence and read out its answer, showing when it holds its private evidence and when it copies. Editing an agent's crop traces planted evidence through the swarm. As in activation patching \citep{meng2022locating,zhang2024patching}, we replace one input, hold everything else fixed, and follow the change through to the collective outcome. Together they show what an agent does with what it hears, and how far what one agent says can travel.

\begin{figure}
    \centering
    \includegraphics[width=0.95\linewidth]{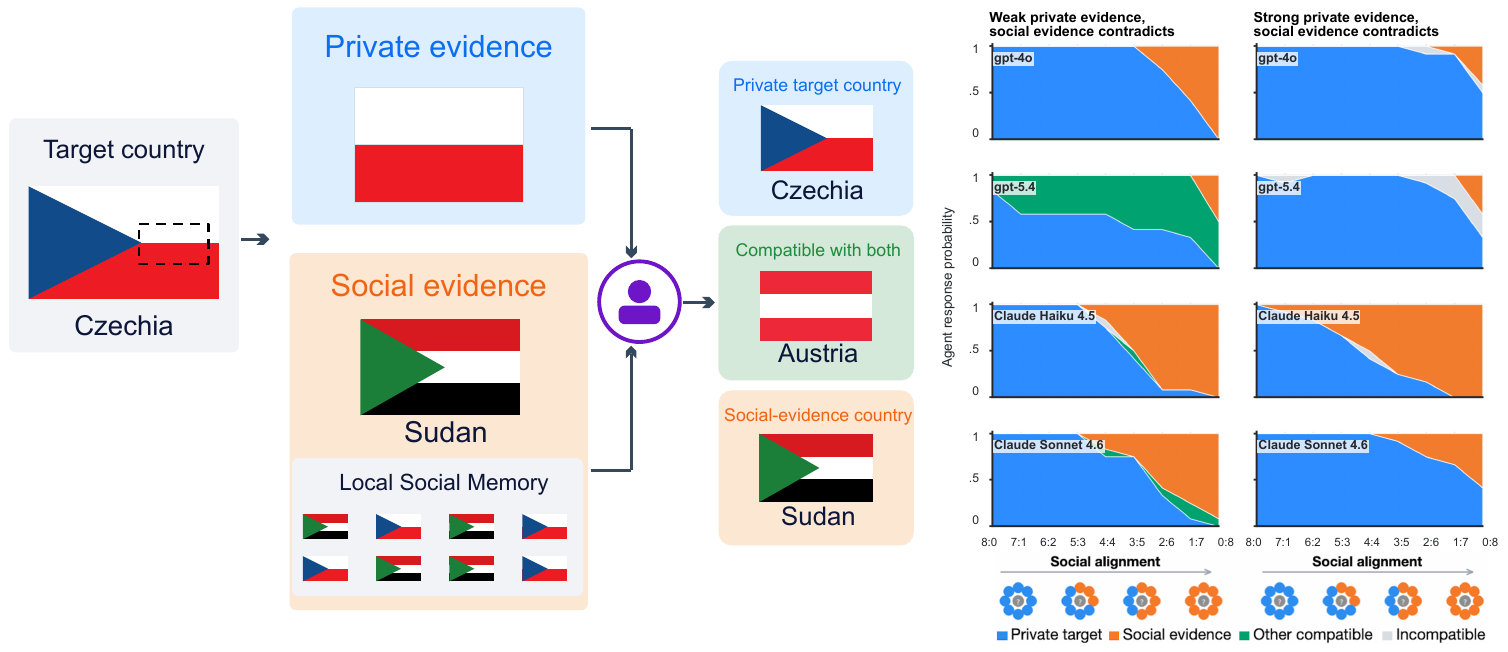}
    \caption{\textbf{Local social memory probe isolates update behavior conditional on private evidence.} (a) A target flag gives private evidence to the agent, while memory contains a shuffled mixture of target country and conflicting social country entries. (b) Agent responses as the memory ratio shifts from target-heavy to social-heavy, across two private-evidence regimes: weak (left), where multiple countries match the private crop, and strong (right), where only the target country matches.}

    \label{fig:memory-conflict-probe}
\end{figure}

\textbf{Local social memory probing.} Whether an agent resists or follows social evidence is a property of the model as much as of the crop. Fig.~\ref{fig:memory-conflict-probe} tests these update behaviors by fixing a private crop, while the target:social ratio in an agent's memory changes. Under weak private evidence, GPT-5.4 shows the highest rate of compatibility reasoning, routing probability into other countries rather than copying the social label. Under strong private evidence (crop uniquely identifies the target), GPT-4o and GPT-5.4 hold firm, while Claude Haiku 4.5 abandons the private target as social memory accumulates, a signature of sycophantic override. We extend this probe in Appendix~\ref{app:memory-probe} with a control condition where social evidence agrees with the private target. This effect may help explain the team diversity result of Sec.~\ref{sec:empirical}: pairing a literal listener with a compatibility-reasoning listener can identify truth-supporting evidence that neither homogeneous team has, consistent with the gains in Fig.~\ref{fig:diversity}a.

\begin{wrapfigure}{r}{0.53\linewidth}
    \vspace{-\baselineskip}
    \centering
    \includegraphics[width=\linewidth]{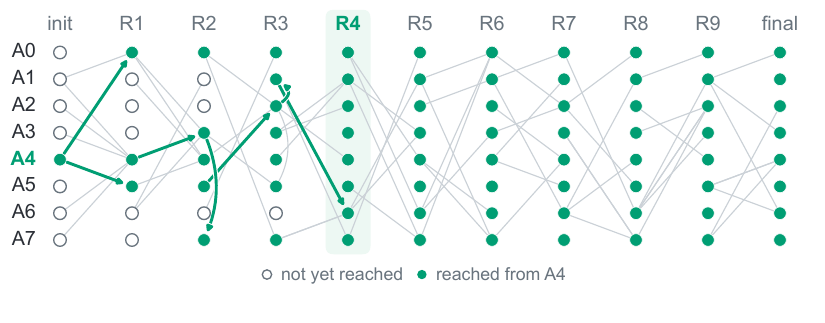}
    \caption{\textbf{Social circuit attribution.} Rows are agents, columns are rounds, and green lines are earliest-arrival paths through the communication schedule. A line within a column marks an agent that received the information and passed it on within the same round. Nodes fill green once A4's message has reached an agent, which happens for every peer by R4.}
    \label{fig:social-circuit}
    \vspace{-\baselineskip}
\end{wrapfigure}

\begin{figure}
    \centering
    \includegraphics[width=1.0\linewidth]{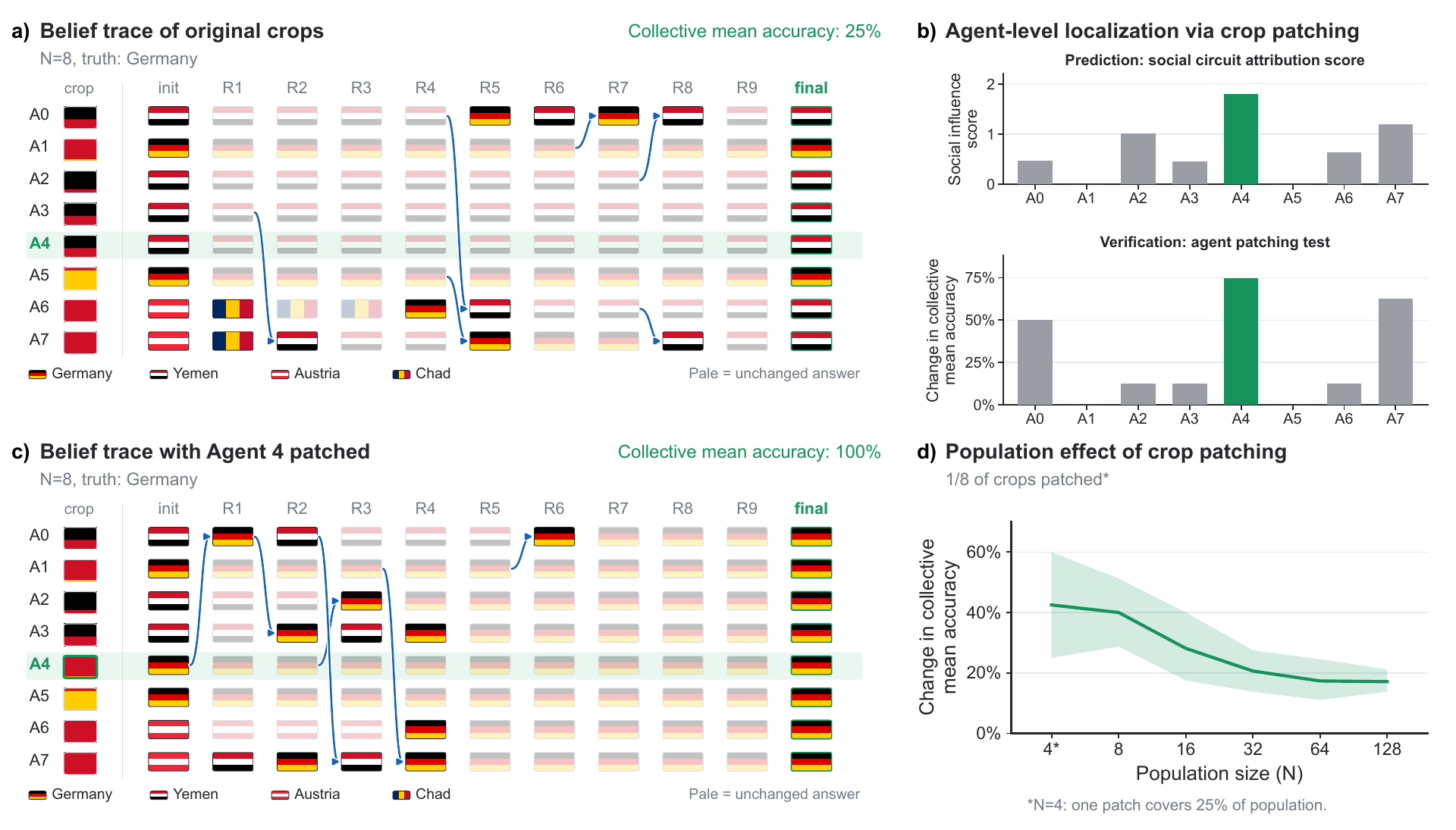}
    \caption{\textbf{Social circuit attribution and its causal verification by agent patching.} (a) Belief trajectories for a \(N=8\) Germany game using original crops. (b) Predicted influence scores (top) and empirical test patching each agent’s crop separately (bottom) both identify Agent 4 as producing the largest collective accuracy improvement. (c) Belief trajectories under the same communication schedule with A4 patched. (d) Change in collective mean accuracy with Agent 4's new crop, averaged across ten runs. One crop patched at \(N=4\), and one-eighth of crops at \(N\geq8\).}

    \label{fig:crop-patching}
\end{figure}

\textbf{Social circuit attribution.}
Using Germany's flag, we choose one informative crop that provides a consistent cue where over ten isolated single-agent probes without social input, it elicited Germany in 10/10 judgments. Before patching this crop into any agent, we quantify each agent's predicted influence by combining the crop's accuracy gain \(\Delta p_i\) (informative crop accuracy based on ten probes - initial accuracy based on ten probes) with its temporal closeness \(E_i\) \citep{pan2011temporal}, how quickly information could reach other agents, directly or through intermediaries. Their product, \(S_i = \Delta p_i E_i\), is the social circuit attribution score. Although A0, A3 and A4 have identical original crops, and hence identical \(\Delta p_i\), their positions in the social circuit differ based on the communication schedule (Fig.~\ref{fig:social-circuit}), and their resulting influence score ranks A4 highest (Fig.~\ref{fig:crop-patching}b, top). We describe our method for ranking these transmission routes, and the assumptions involved, in Appendix~\ref{app:temporal-intervention-sites}.

\textbf{Verification by agent patching.}
We then empirically test the intervention our influence score ranks. We patch each agent's crop separately with the same informative crop, holding the communication schedule fixed, and measure the change in final collective mean accuracy. At \(N=8\), patching A4 produces the largest observed improvement (Fig.~\ref{fig:crop-patching}b, bottom), matching our method's ranking.

\textbf{Agent-level causal tracing via crop patching.}
The selected paired traces illustrate the result: collective mean accuracy increases from 25\% with original crops to 100\% with A4 patched (Fig.~\ref{fig:crop-patching}a,c). Arrows mark country switches consistent with the latest received message, showing routes along which the patch may have spread. We then repeat the comparison across ten runs, evaluating collective mean accuracy after ten rounds. For \(N=8-128\), we patch the same proportion of agents (\(1/8\)). The mean improvement decreases from 40\% at \(N=8\) to approximately 17\% at \(N=128\) (Fig.~\ref{fig:crop-patching}d). Thus, the same intervention fraction produces less collective correction at larger populations.

\textbf{From agents to populations.} Fig.~\ref{fig:crop-patching}d shows how a fixed share of planted evidence has a smaller impact on collective performance as \(N\) grows. At small \(N\), the tools from mechanistic interpretability suffice, since a patch on one agent shifts the outcome and an interaction trace shows how. As N grows, the same patch has a smaller effect and the swarm enters a regime where collective belief is a property of the population rather than of the agents in it. That regime calls for a statistical mechanical view, in which collective belief follows from the composition and size of the population rather than any individual members. We develop this next and return with it to the population sweep of Fig.~\ref{fig:population_sweep}b.

\subsection[Statistical mechanics of bounded agents]{Statistical mechanics of bounded agents}
\label{sec:stat-mech}

Here we present a simple phenomenological theory that captures three key features of the experiment: as population size increases, collective belief collapse decreases, collective belief polarization increases, and collective performance is highest at an intermediate population size. With a small set of assumptions, we aim to provide an intuitive physical picture of these phenomena. We start by adding private evidence from the external world to Quantized Simplex Gossip \citep{tanaka2026lottery}.

\begin{figure}
    \centering
    \includegraphics[width=1\linewidth]{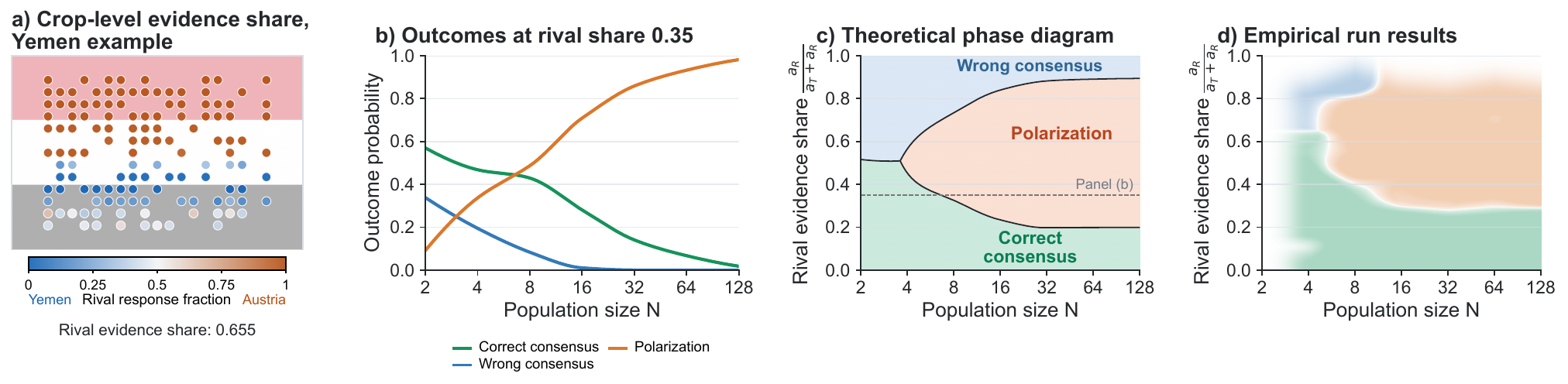}
    \caption{\textbf{Private evidence and collective outcomes across population size.} (a) Rival response fractions from 50 binary probes per crop in an $N=128$ Yemen run: blue supports Yemen, orange supports Austria. (b) Outcome probabilities at a rival evidence share $=0.35$ with $a_T+a_R=0.45$ and $h_0=+0.3$. (c) Theoretical phase diagram of population size and rival evidence share. (d) Empirical run results from population sweep, using each run's probed rival evidence share. Regions show the locally most frequent outcome, estimated with Gaussian bandwidth $0.05$ in share.}
    \label{fig:phase-diagram}
\end{figure}

\textbf{Private evidence.} We simplify the multi-country experiment to two beliefs, the truth country \(T\) and one rival country \(R\). The same framework we describe below can extend to accommodate additional labels. A crop can favor either belief or leave the agent uncertain, as in Fig.~\ref{fig:phase-diagram}a's Yemen--Austria example, black distinguishes Yemen from Austria, a red-and-white crop can be read as Austria, and an ambiguous crop leaves many possibilities open. We describe crops as three types: truth-deciding, rival-deciding, and ambiguous. Each agent independently draws a type with probabilities \(a_T\), \(a_R\), and \(a_0=1-a_T-a_R\). These are probabilities under crop sampling, rather than literal fractions of the flag's area. To compare this with our multi-country experiments, we probe each initial crop used in our population sweep 50 times restricting country answers to \(T\) and \(R\). The fraction of valid responses choosing \(R\), pooled across crops, sets the rival evidence share \(a_R/(a_T+a_R)\).

\textbf{Social updating.} Agents with truth- or rival-deciding evidence mainly keep their initial belief. They are evidence-induced \emph{zealots}, or agents whose private evidence resists social pressure. Ambiguous agents instead copy a randomly chosen speaker, adopting a heard \(T\) with probability \(q_T\) and a heard \(R\) with probability \(q_R=1-q_T\); otherwise they retain their current belief. We write \(q_T/q_R=e^{h_0}\). When \(h_0=0\), this is neutral copying, with an equal probability of accepting either label. When \(h_0<0\), ambiguous agents lean toward the rival at each update. This bias represents an interpretation tendency under uncertain evidence; here it is a modeling assumption. The model extends the neutral copying limit of QSG \citep{tanaka2026lottery} with evidence-induced zealots \citep{mobilia2003zealot,mobilia2007zealotry} and biased adoption. With no zealots and \(h_0=0\), accepted updates between distinct agents follow the binary \(\alpha=m=1\) copying limit of QSG.

\textbf{Microscopic dynamics.} In a population of \(N\) agents, let \(z_T\) count truth zealots, \(z_R\) rival zealots, and \(z_0=N-z_T-z_R\) ambiguous agents. Their population fractions are \(f_X=z_X/N\), for \(X\in\{T,R,0\}\). Thus \(a_X\) is a sampling probability, while \(f_X\) is the fraction actually sampled in one population. Let \(n\) count ambiguous agents that currently identify the flag as \(T\). Each update samples a speaker and listener independently and uniformly. The probabilities that \(n\) increases or decreases by one are
\begin{equation}
W_+(n)=q_T\frac{(z_0-n)(z_T+n)}{N^2},\qquad
W_-(n)=q_R\frac{n(z_R+z_0-n)}{N^2}.
\label{eq:rates}
\end{equation}
For example, \(W_+\) is the probability of choosing an ambiguous listener who currently identifies the flag as \(R\), a truth-holder as speaker, and accepting the message. The remaining probability gives no change. When \(z_0>0\), \(x=n/z_0\) is the truth fraction among ambiguous agents, and \(s=f_T+f_0x\) is the truth fraction in the whole population. If \(z_0=0\), the population is fixed at \(s=f_T\).

\textbf{Drift and fluctuations.} For \(\tau=t/N\), Eq.~\eqref{eq:rates} gives the mean-field drift
\begin{equation}
\frac{dx}{d\tau}=A(x)
=q_Tf_T(1-x)-q_Rf_Rx+(q_T-q_R)f_0x(1-x).
\label{eq:meanfield}
\end{equation}
The first term increases \(x\) when ambiguous agents holding \(R\) copy truth zealots. The second decreases \(x\) when ambiguous agents holding \(T\) copy rival zealots. The third describes copying between ambiguous agents: it vanishes under neutral copying and favors truth when \(h_0>0\). Over a fixed interval of \(\tau\), copying fluctuations scale as \(N^{-1/2}\) at fixed type fractions with \(f_0>0\) (Appendix~\ref{app:minimal-theory}). Larger populations therefore follow the mean-field dynamics more closely, but finite populations can continue changing even when \(A(x)=0\).

\textbf{Evidence coverage.} At fixed \(a_T,a_R\), increasing \(N\) samples more evidence without changing its underlying distribution. The probabilities of missing each zealot type are
\begin{equation}
    \Pr(z_T=0)=(1-a_T)^N,\qquad
    \Pr(z_R=0)=(1-a_R)^N.
    \label{eq:evidence-absence}
\end{equation}
The characteristic population sizes for encountering these types are therefore \(Na_T\sim1\) and \(Na_R\sim1\). One can picture \(1/N\) as a waterline covering the sampling probabilities \(a_T\) and \(a_R\).  As \(N\) increases, the water recedes, illustrating how commonly sampled evidence tends to appear before rarer evidence. When \(a_T>a_R\), truth evidence therefore tends to appear first. These scales organize three overlapping population phases, whose boundaries are finite-population crossovers. Fig.~\ref{fig:phase-diagram}b--c shows the endpoint probabilities from Eq.~\eqref{eq:rates} and regions identifying the most probable collective outcome. We use \(a_T+a_R=0.45\) and \(h_0=0.3\), which favors truth adoption, chosen for qualitative agreement with the empirical population trends. The binary probes described above set the relative evidence share.

\textbf{Memetic-drift phase.} At small \(N\), many populations contain no zealots. Setting \(f_T=f_R=0\) and \(h_0=0\) in Eq.~\eqref{eq:meanfield} gives \(A(x)=0\) where neither label has a deterministic advantage. Copying fluctuations nevertheless lead each finite population to consensus by chance, the memetic-drift mechanism of neutral QSG. Truth-biased copying reduces, but does not eliminate, the chance of wrong consensus. Small populations containing rival zealots but no truth zealots instead reach wrong consensus through persistent rival evidence. Both routes can produce collective belief collapse.

\textbf{Wisdom-of-crowds phase.} At intermediate \(N\), truth-deciding agents are increasingly present while rival-deciding agents are still scarce. With \(f_R=0\), Eq.~\eqref{eq:meanfield} has the fixed point \(x^\ast=1\), which is locally stable when \(q_Tf_T>(q_R-q_T)f_0\). Without rival zealots, a finite population containing truth zealots eventually reaches correct consensus: communication spreads the evidence of a few agents to the rest. Such a window requires more than \(a_T>a_R\) alone and need not produce a peak in correct-consensus probability. Finite populations with truth zealots only eventually reach truth consensus even outside the deterministic stability condition, but the waiting time can be long (Appendix~\ref{app:minimal-theory}).

\textbf{Polarization phase.} At larger \(N\), both kinds of zealots are typically present when \(a_T,a_R>0\). For \(f_0>0\), \(A(0)>0\) and \(A(1)<0\), neither unanimous state is sustained, and the mean-field dynamics has a stable interior fixed point. Under neutral copying, \(x^\ast=f_T/(f_T+f_R)\); rival-biased adoption shifts it downward. Neither zealot type can convert the other, and their repeated messages sustain competing beliefs among ambiguous agents. This split keeps the population away from perfect convergence but retains some of the truth within the collective. At fixed positive evidence fractions, the fluctuations shrink with \(N\), making the split more persistent. Whether a particular split meets the empirical polarization thresholds also depends on the relative sizes of the two camps. A truth-dominated split may still count as correct consensus, so the green region need not vanish at large populations.

\textbf{Accuracy peak.} With fixed \(a_T>a_R\), neutral copying increases mean truth share monotonically with \(N\), so evidence coverage alone does not explain the mean accuracy peak in this model. In populations with more truth zealots than rival zealots, truth can still spread widely; when both types are similarly represented, the same bias shifts ambiguous agents toward the rival and lowers the truth share. Averaging model expectations at each run’s measured evidence share reproduces an intermediate-population accuracy peak (Appendix~\ref{app:minimal-theory}). The model thus separates two roles of private evidence: it can correct an arbitrary consensus, and it can hold a locally plausible false belief in place.

\textbf{Complementary approaches.} Causal interventions on agents and statistical mechanics rest on opposite assumptions for how a swarm is organized. Mechanistic interpretability of neural networks relies on causal interventions as optimization of weights gives rise to emergent representational structures, where concepts can be localized in a low-dimensional subspace and then ablated or patched. Statistical mechanics instead often assumes random, unstructured connections, as in the theory of randomly connected neurons, and describes the competition between noise and bias amplification. The swarms in the incidents above built message boards that centralized their connections, which is where causal interventions on individual agents are informative. But their large-scale communication structures formed spontaneously, motivating a complementary statistical-mechanical description.

\FloatBarrier
\section{Conclusion}

The goal of this work was to introduce a toy model and develop principled frameworks to make scientific progress on mechanistic swarm interpretability. The Flag Game is a steerable system that captures key features of real-world AI swarm coordination. Using the Flag Game, we found that collective belief collapse crosses over to collective belief polarization as the number of agents increases. Although polarization lowers mean accuracy, it can be a less dangerous failure mode than belief collapse, as a population that collapses onto a false belief has nothing left with which to correct itself, whereas a polarized population retains competing beliefs. In that sense, polarization can be a first step toward plurality. For safety purposes, what to watch out for may therefore not be disagreement, but consensus of agents’ beliefs or intent under social pressure.

Overall, the recent emergence of a sociology of AIs suggests that aligning AI swarms may require a different paradigm from aligning an individual AI agent. Just as qualitatively different mechanisms emerge in collections of atoms, collective systems of agents may require new variables and new laws of description. The state we want to achieve may itself need to be specified collectively, for example, as a plural state characterized by diversity of beliefs rather than consensus under social pressure. Mechanistic swarm interpretability should therefore identify how the microscopic properties of agents and their communication give rise to macroscopic collective states, ultimately enabling the inverse design of swarms toward desired collective behavior.

Our complementary toolbox of causal interventions and statistical-mechanical approaches may be useful not only across population sizes, but also across time. At small population sizes, causal interventions can identify which agent, memory, or piece of evidence matters to a collective outcome. In the recent OpenAI/Hugging Face incident, key beliefs and coordination structures emerged while the population was still relatively small, before many more agents joined the swarm. Early in such dynamics, social circuit attribution and agent patching may therefore be most informative. As the population grows, however, the same local intervention produces less collective correction, and population-level variables such as evidence coverage, communication structure, and collective order parameters may become more useful. Mechanistic swarm interpretability may therefore require a scale-adaptive approach: causal interventions to reverse engineer social circuits during early swarm formation, and statistical mechanics to understand and control the collective phases that emerge as the swarm grows. As intelligence scales from networks of neurons to larger networks of interacting agents, we hope our empirical and theoretical frameworks can help us climb the ladder of complexity with clarity.

\bibliographystyle{unsrtnat}
\bibliography{main}

\newpage
\appendix

\section{Limitations}
\label{app:limitations}

The Flag Game is a controlled diagnostic task rather than a full model of real-world tasks. Insights gained through it about coordination and social influence in multi-agent systems may be over-interpreted as evidence of real-world scenario reliability, or potentially misused to design interventions that steer group decisions. We release only synthetic task and evaluation code, and real-world multi-agent systems require additional domain-specific, safety, and human-centered evaluation. Its simplicity makes the social mechanisms measurable, but trends will depend on the chosen data set, prompts, memory format, communication protocols, and aggregation rules. In particular, prompt wording is an important experimental variable, as different instructions about how to use social evidence could change the degree of belief collapse or polarization observed.

Our model comparisons also cover only a slice of the space of possible model identities. The two main models tested already exhibited substantial heterogeneity, which was sufficient to reveal composition and manager effects, but the paper does not claim to characterize model diversity exhaustively. Finally, exhaustive population sweeps scale with the number of agents, trials, rounds, protocols, and model calls, making broader searches over architectures and model mixtures computationally expensive. The paper's work compute was approximately \$25,000 total hosted API cost, main runs used GPT-4o and GPT-5.4 and smaller validation runs used Claude Haiku 4.5 and Claude Sonnet 4.6.

\section{Pairwise alpha sweep}

\label{app:alpha-sweep}
\begin{figure}[ht]
    \centering
    \includegraphics[width=0.55\linewidth]{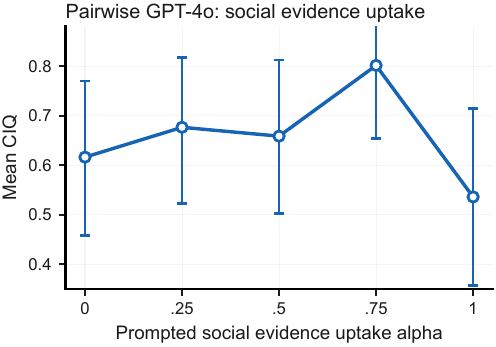}
    \caption{The pairwise protocol has an interior optimum at 0.75, showing that in local exchange higher uptake can create overdependence on incorrect speakers.}
    \label{fig:alpha-sweep-appendix}
\end{figure}

\section{Full single agent memory probe}
\label{app:memory-probe}

\begin{figure}[ht]
    \centering
    \includegraphics[width=0.8\linewidth]{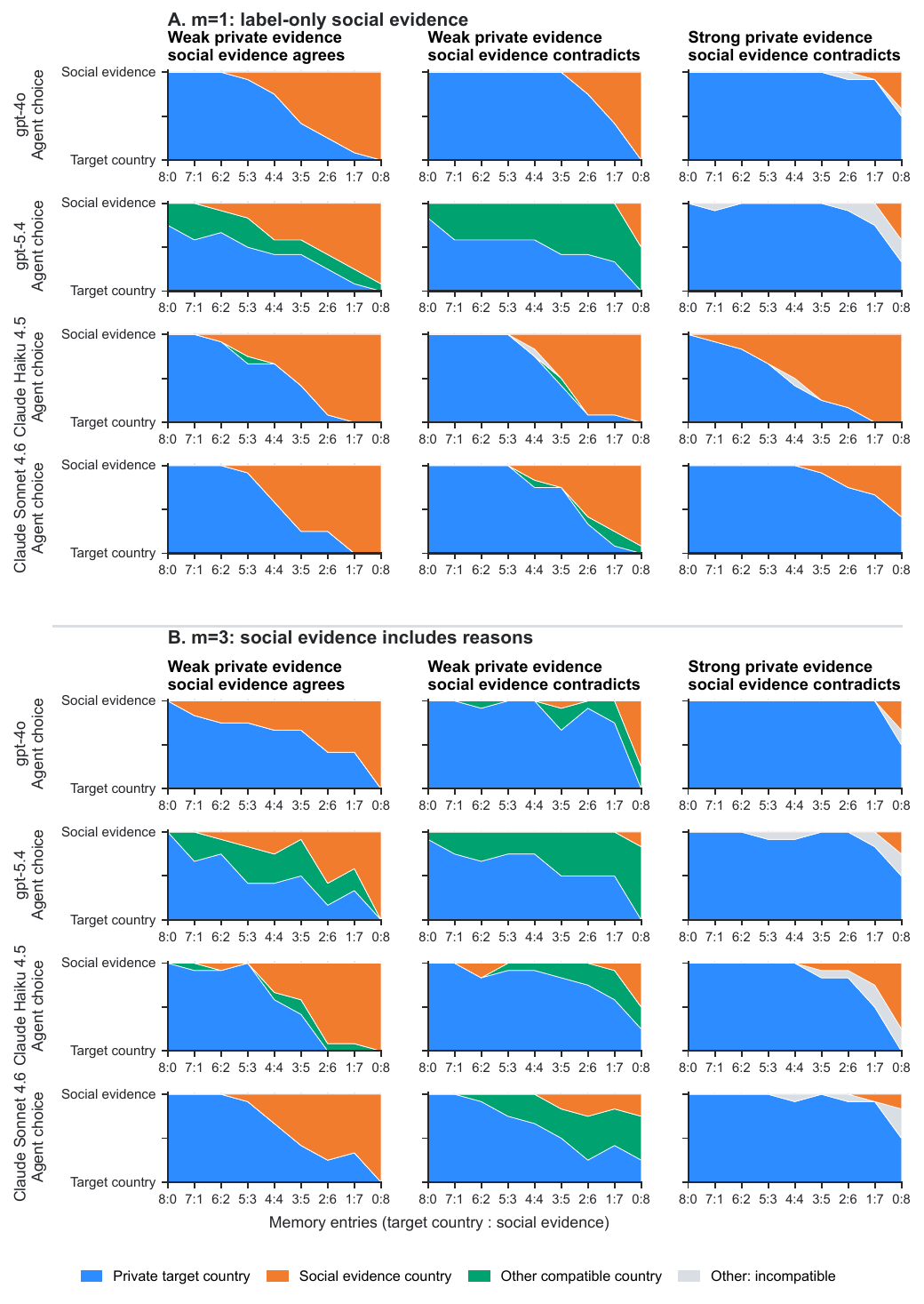}
    \caption{\textbf{Memory-conflict probe across four models.} Agent responses as local social memory shifts from target-heavy (8:0) to social-heavy (0:8), across three private-evidence regimes and two message bandwidths (m=1 label-only; m=3 with reasons).}
    \label{fig:memory-extension}
\end{figure}

Fig.~\ref{fig:memory-extension} extends the memory-conflict probe in Fig.~\ref{fig:memory-conflict-probe} to the following four models: GPT-4o, GPT-5.4, Claude Haiku 4.5, and Claude Sonnet 4.6, and three regimes: weak private evidence (indicating multiple countries are compatible with the crop) and compatible social evidence, weak private evidence and incompatible social evidence, and strong private evidence (indicating only the target country is compatible with the crop) and incompatible social evidence. GPT-5.4 routes substantially more probability mass into other crop-compatible countries (green), whereas GPT-4o, Claude Haiku 4.5, and Claude Sonnet 4.6 update more literally between the private target and the social-evidence country. At m=3, when social memory includes reasoning, Haiku and Sonnet also start to respond with compatible alternatives, but the effect remains strongest in GPT-5.4. Strong private evidence (rightmost column) shows that GPT-4o, GPT-5.4, and Sonnet hold firm on the target, whereas Haiku is more readily moved by social-heavy memory even with strong private evidence.

\section{Empirical run details}
\label{app:empirical-run-details}

\begin{table}[ht]
    \centering
    \small
    \caption{Run settings behind the empirical panels in Sec.~\ref{sec:empirical}.}
    \label{tab:empirical-run-families}
    \begin{tabular}{p{0.24\linewidth}p{0.16\linewidth}p{0.31\linewidth}p{0.16\linewidth}}
        \toprule
        Figure slice & Protocol & Main controls & Trials used \\
        \midrule
        Population scaling & Pairwise & \(m=3\); no social-awareness prompt & 40 seeds per (model, N) \\
        Protocol side-by-side & Pairwise, broadcast, manager & \(N=8\); \(m=3\); no social-awareness prompt & 60 matched seeds per condition \\
        Pairwise social-awareness & Pairwise & GPT-4o; \(N=16\); \(m=3\) & 28 seeds per $\alpha$ \\
        Broadcast social-awareness and composition & Broadcast & \(N=8\); \(m=3\) & 30 seeds per \((\alpha,\mathrm{composition})\) \\
        Memory-conflict probe & Single-agent prompt probe & GPT-4o, GPT-5.4, Claude Haiku 4.5 and Sonnet 4.6; \(m\in\{1,3\}\) & 36 trials \\
        \bottomrule
    \end{tabular}
\end{table}

\begin{table}[t]
\centering
\small
\setlength{\tabcolsep}{5pt}
\caption{Endpoint-threshold robustness on the pairwise communication protocol for GPT-4o.}
\label{tab:flag-threshold-robustness}
\begin{tabular}{rrrrr}
\toprule
$N$ & Correct cons. (\%) & Wrong cons. (\%) & Polarized (\%) & Fragmented (\%) \\
\midrule
4 & 50.0--52.6 & 26.3--34.2 & 13.2--23.7 & 0.0--10.5 \\
8 & 47.4 & 5.3--23.7 & 26.3--44.7 & 0.0--21.1 \\
16 & 50.0--52.6 & 0.0--10.5 & 18.4--44.7 & 0.0--31.6 \\
32 & 42.1--47.4 & 0.0 & 31.6--55.3 & 0.0--26.3 \\
64 & 36.8--42.1 & 0.0--5.3 & 34.2--55.3 & 0.0--28.9 \\
128 & 37.8--40.5 & 0.0 & 56.8--59.5 & 0.0--5.4 \\
\bottomrule
\end{tabular}
\end{table}

Here in Table~\ref{tab:empirical-run-families}, we record the run settings behind the empirical panels in Sec.~\ref{sec:empirical}. All reported runs use temperature \(0.2\), top-p \(1.0\), and a multimodal user message consisting of the text prompt plus the private crop image with high image detail. Unless stated otherwise, images are rendered on a \(24\times 16\) canvas and agents receive \(6\times 4\) crops at render scale 25. Memory buffers store at most \(H=8\) prior entries per agent. Pairwise calls use a 200-token completion cap in the population and social-awareness sweeps; broadcast and manager calls use a 250-token cap. Each trial samples a hidden country uniformly from the country pool of 28 stripe and triangle flags. Additionally, in Table~\ref{tab:flag-threshold-robustness} we show the robustness of our endpoint results with different thresholds used for consensus and polarization. Entries give minimum--maximum percentages over 30 different combinations of consensus thresholds $0.75$--$1.00$ and polarization thresholds $0.15$--$0.35$, each in steps of $0.05$. Ranges measure threshold sensitivity, not statistical uncertainty. The values we settled on for our analysis are $0.85$ for consensus, both correct and wrong, and $0.25$ for polarization.

\paragraph{Model composition.}
Mixed \(N=8\) conditions assign four observer agents to GPT-5.4 and four to GPT-4o. In the broadcast composition sweep, the GPT-5.4 count ranges from 0 to 8 while the remaining agents are GPT-4o. Manager conditions have \(N=8\) crop-bearing observers plus one blind manager synthesizer. The manager slot is either GPT-4o or GPT-5.4; the observer group is all GPT-4o, all GPT-5.4, or a 4/4 mix.

\section{Prompting and social-awareness intervention}
\label{app:prompt-details}

All protocol prompts specify a JSON-only answer and attach the private crop image to the same user message (unless the agent is a manager). The message bandwidth \(m\) determines the JSON schema: \(m=1\) asks only for a country and \(m=3\) asks for a country and one-sentence reason.

\begin{figure}[htbp]
\centering
\begin{minipage}{\linewidth}

\begin{promptbox}{System message}
You must output only valid JSON. No extra keys, no markdown,
and no text outside the JSON object.
You are one player in a flag identification game.
Choose exactly one country from the allowed countries listed in the user message.
Follow the exact output schema given in the user message.
\end{promptbox}

\begin{promptbox}{User message}
All players are identifying the same underlying flag.
You always see the same private crop.
Transcript memory shows messages you observed from previous interactions with other players.

Allowed countries: <JSON list>
Transcript memory (oldest -> newest):
<speaker messages, or []>
<optional social-awareness line>

Output JSON exactly:
{"country":"<one allowed country>","reason":"<one sentence>"}
\end{promptbox}

\end{minipage}

\caption{\textbf{Example pairwise prompt.}
The pairwise protocol uses the same text for interaction messages and probe queries, differing only in the schema line.}
\label{fig:pairwise-prompt}
\end{figure}

\paragraph{Social-awareness line.}
When \texttt{prompt\_social\_susceptibility} is enabled, the prompt inserts guidance based on $\alpha$ (Table~\ref{tab:susceptibility-prompt-ladder}).

\begin{table}[ht]
    \centering
    \small
    \caption{Social-awareness prompt ladder indexed by $\alpha$.}
    \label{tab:susceptibility-prompt-ladder}
    \begin{tabular}{p{0.12\linewidth}p{0.40\linewidth}p{0.40\linewidth}}
        \toprule
        Range & Pairwise wording & Broadcast wording \\
        \midrule
        \(\alpha\leq .2\) & Rely mostly on your own crop and treat transcript memory as weak evidence. & Rely mostly on your own evidence; treat other agents' country guesses as weak evidence. \\
        \(.2<\alpha\leq .4\) & Give somewhat more weight to your own crop than to transcript memory. & Give somewhat more weight to your own evidence than to other agents' country guesses. \\
        \(.4<\alpha\leq .6\) & Balance your own crop and transcript memory. & Balance your own evidence with other agents' country guesses, using their guesses as real evidence. \\
        \(.6<\alpha\leq .8\) & Give somewhat more weight to transcript memory than to your own crop. & Give somewhat more weight to other agents' country guesses than to your own evidence. \\
        \(\alpha>.8\) & Treat transcript memory as strong evidence and update readily toward it. & Treat other agents' country guesses as strong evidence and update readily toward them. \\
        \bottomrule
    \end{tabular}
\end{table}

\paragraph{Memory-conflict probe prompt.}
The memory-conflict probe reuses the pairwise prompt with no live social interaction. For each trial, the agent receives one crop from a target country and a synthetic memory of eight entries. If \(k\) is the false-memory count, then \(k\) memory entries name a lure country and \(8-k\) entries name the target country, shuffled before prompting.

\section{Temporal communication analysis}
\label{app:temporal-intervention-sites}

We summarize opportunities for information to spread using time-ordered communication paths \citep{pan2011temporal}. Let \(\tau_{ij}(0)\) be the earliest global message index at which information starting at agent \(i\) at initialization could reach agent \(j\), directly or through intermediaries. Contacts must occur in chronological order, with information allowed to wait between them. We report mean arrival time $D_i$ and temporal closeness $E_i$, the mean inverse arrival time:
\[
D_i = \frac{1}{N-1}\sum_{j\ne i}\tau_{ij}(0),
\qquad
E_i = \frac{1}{N-1}\sum_{j\ne i}\frac{N}{\tau_{ij}(0)}.
\]
$D_i$ is measured in interactions; dividing arrival times by $N$ expresses them in rounds before taking their reciprocals in $E_i$. The latter follows the mean-inverse form of temporal closeness and efficiency, evaluated from initialization. Each recipient contributes once, according to its earliest arrival, so earlier access contributes more. An unreached recipient would contribute zero to $E_i$.

We combine this schedule-based measure with the accuracy gain from replacing agent $i$'s crop:
\[
\Delta p_i =
\hat p_{\mathrm{informative}}-\hat p_{\mathrm{original},i},
\qquad
S_i = \Delta p_i E_i.
\]
Each original-crop and informative crop accuracy is estimated from ten probes. Thus, the social circuit attribution score $S_i$ combines the additional local evidence of the patch with opportunities for its early dissemination. Table~\ref{tab:temporal-influence} breaks down the calculation across all agents, showing A4 has the highest predicted influence score and subsequently the largest observed patching effect.

\begin{table}[t]
\centering
\small
\begin{tabular}{lrrrrrr}
\toprule
Agent & Original correct & $\Delta p_i$ & $D_i$
      & $E_i$ & $S_i$ & $\Delta A_i$ \\
      & (out of 10) & & & & & \\
\midrule
A0 &  0 & 1 & 19.4 & 0.47 & 0.47 & 0.50 \\
A1 & 10 & 0 & 14.9 & 0.73 & 0.00 & 0.00 \\
A2 &  0 & 1 & 16.4 & 1.01 & 1.01 & 0.13 \\
A3 &  0 & 1 & 22.9 & 0.46 & 0.46 & 0.13 \\
A4 &  0 & 1 & 12.9 & 1.81 & \textbf{1.81} & \textbf{0.75} \\
A5 & 10 & 0 & 24.6 & 0.39 & 0.00 & 0.00 \\
A6 &  0 & 1 & 15.6 & 0.64 & 0.64 & 0.13 \\
A7 &  0 & 1 & 16.6 & 1.19 & 1.19 & 0.63 \\
\bottomrule
\end{tabular}
\caption{Agent-level metrics for the selected communication schedule. $\Delta A_i$ is the empirical change in collective mean accuracy relative to the original-crop baseline of 25\%, from one game per patch.}
\label{tab:temporal-influence}
\end{table}

\section{Minimal theory: finite-population details}
\label{app:minimal-theory}

\textbf{Evidence composition.} Let \((z_T,z_R,z_0)\sim\mathrm{Multinomial}(N;a_T,a_R,a_0)\), where \(a_0=1-a_T-a_R\). The four mutually exclusive evidence compositions have exact probabilities
\begin{align}
P_{\mathrm{neither}}&=a_0^N,\qquad
P_{T\text{-only}}=(1-a_R)^N-a_0^N,\nonumber\\
P_{R\text{-only}}&=(1-a_T)^N-a_0^N,\nonumber\\
P_{T,R}&=1-(1-a_T)^N-(1-a_R)^N+a_0^N.
\label{eq:evidence-compositions}
\end{align}
The probability of having no truth zealot combines unanchored populations and populations with rival zealots only, with total probability \((1-a_T)^N\). The scales \(Na_T\sim1\) and \(Na_R\sim1\) indicate when the corresponding evidence starts to appear; they are crossover estimates, not sharp boundaries or definitions of empirical endpoint classes.

\paragraph{Exact finite-population dynamics.} The transition probabilities in Eq.~\eqref{eq:rates} define the process conditional on \((z_T,z_R,z_0)\). Let \(p_t(n)\) be the probability that \(n\) ambiguous agents report truth after \(t\) interaction steps. Then
\begin{align}
 p_{t+1}(n)-p_t(n)
 ={}&W_+(n-1)p_t(n-1)+W_-(n+1)p_t(n+1)\nonumber\\
 &-[W_+(n)+W_-(n)]p_t(n).
 \label{eq:master-equation}
\end{align}
Terms outside \(0\leq n\leq z_0\) are zero. Selecting the same agent as speaker and listener, or rejecting a message, leaves the state unchanged. The theoretical panels in Fig.~\ref{fig:phase-diagram} use the stationary and fixation probabilities of this process, with \(q_T=(1+e^{-h_0})^{-1}\), averaged over the sampled evidence compositions.

\paragraph{Mean-field closure.}
For \(z_0>0\), the exact one-step mean satisfies
\(\mathbb E[x_{t+1}]-\mathbb E[x_t]=\mathbb E[A(x_t)]/N\), conditional on the evidence composition. Replacing \(\mathbb E[A(x)]\) by \(A(\mathbb E[x])\) gives Eq.~\eqref{eq:meanfield}; when \(h_0\ne0\), this is an approximation because \(A\) is nonlinear. Writing \(\Delta x=x_{t+1}-x_t\), the one-step variance is
\begin{equation}
 \operatorname{Var}(\Delta x\mid n)
 =\frac{W_++W_--(W_+-W_-)^2}{z_0^2}.
 \label{eq:update-variance}
\end{equation}
The drift and fluctuations therefore follow from the same update rule. At fixed type fractions with \(f_0>0\), fluctuations over a fixed interval of \(\tau=t/N\) scale as \(N^{-1/2}\)

\paragraph{Memetic drift and truth spreading.}
Without zealots, the probability of eventual truth consensus from \(n\) initial truth reports is
\begin{equation}
 \Pr(T\text{ fixes}\mid n)=
 \begin{cases}
 n/N,&h_0=0,\\[3pt]
 \displaystyle\frac{1-e^{-h_0n}}{1-e^{-h_0N}},&h_0\ne0.
 \end{cases}
 \label{eq:unanchored-fixation}
\end{equation}
Under neutral copying, finite populations therefore reach consensus by chance despite \(A(x)=0\). Positive bias raises the truth-fixation probability. We average over independent fair initial beliefs, \(n\sim\operatorname{Binomial}(N,1/2)\).

For \(0<q_T,q_R<1\), a finite population with truth zealots and no rival zealots eventually reaches truth consensus. In the truth-only case, \(A'(1)=-q_Tf_T+(q_R-q_T)f_0\), so truth consensus is locally stable when \(q_Tf_T>(q_R-q_T)f_0\). If this inequality is strictly reversed, the mean-field dynamics can settle at a mixed state. Random fluctuations still eventually bring the finite population to truth consensus, but this may take longer than the available interaction budget.

\paragraph{Mean accuracy.}
Under neutral copying, populations with zealots have expected truth share \(z_T/(z_T+z_R)\). Averaging over compositions with unbiased initial beliefs  gives:
\begin{equation}
 \mathbb E[s_\infty]
 =\frac{a_T}{a_T+a_R}(1-a_0^N)+\frac12a_0^N.
 \label{eq:neutral-mean-accuracy}
\end{equation}
For fixed \(a_T>a_R\), this is nondecreasing with \(N\): evidence coverage alone does not produce a mean-accuracy peak in the neutral model. Mean truth share differs from correct-consensus probability. Figure~\ref{fig:phase-diagram} uses truth-biased copying, \(h_0=+0.3\), and its fixed-share slice likewise has no mean-accuracy peak. Averaging model expectations at each run's measured share produces a maximum at \(N=32\), similar to the empirical cohort. This comparison averages over the measured evidence shares at each \(N\).

\paragraph{Persistent competition.}
With both zealot types and \(z_0>0\), \(A(0)>0\) and \(A(1)<0\), giving a stable interior fixed point. The exact stationary probabilities satisfy
\begin{equation}
 \frac{\pi_{n+1}}{\pi_n}
 =\frac{W_+(n)}{W_-(n+1)}
 =e^{h_0}\frac{(z_0-n)(z_T+n)}{(n+1)(z_R+z_0-n-1)}.
 \label{eq:stationary-recurrence}
\end{equation}
Normalization determines \(\pi_n\). Under neutral copying this is the beta-binomial law of the zealot voter model \citep{mobilia2007zealotry}, with
\begin{equation}
 \mathbb E[x]=\frac{z_T}{z_T+z_R},\qquad
 \operatorname{Var}(x)
 =\frac{z_Tz_RN}{z_0(z_T+z_R)^2(z_T+z_R+1)}.
 \label{eq:neutral-moments}
\end{equation}
At fixed positive fractions, fluctuations shrink as \(N^{-1/2}\). Positive \(h_0\) tilts the stationary weights toward truth and negative \(h_0\) toward the rival. Both zealot types prevent unanimity, but an unequal split can still meet the consensus threshold. If \(z_0=0\), the truth share remains fixed at \(z_T/N\).

\end{document}